\documentclass{article} %

\usepackage{microtype}
\usepackage{graphicx}
\usepackage{booktabs} %

\usepackage[hyphens]{url}
\usepackage{hyperref}
\usepackage[hyphenbreaks]{breakurl}

\usepackage[preprint]{icml2025}

\usepackage{amsmath}
\usepackage{amssymb}
\usepackage{mathtools}
\usepackage{amsthm}

\usepackage[capitalize,noabbrev]{cleveref}

\theoremstyle{plain}
\newtheorem{theorem}{Theorem}[section]

\theoremstyle{definition}
\newtheorem{definition}[theorem]{Definition}

\theoremstyle{remark}

\usepackage[textsize=tiny]{todonotes}

\usepackage{tabularray}
\UseTblrLibrary{amsmath}

\usepackage{xcolor}
\usepackage[procnames]{listings}

\definecolor{keywords}{RGB}{255,0,90}
\definecolor{comments}{RGB}{0,0,113}
\definecolor{red}{RGB}{160,0,0}
\definecolor{green}{RGB}{0,150,0}
\definecolor{blue}{RGB}{0,0,150}
\usepackage{booktabs}
\usepackage{caption}
\usepackage{subcaption}

\usepackage{pythonhighlight}
\lstnewenvironment{mypython}[1][]{\lstset{#1}}{}

\usepackage{amsmath,amsfonts,bm}

\def\eqref#1{equation~\ref{#1}}

\newcommand{\trace}{\operatorname{tr}}
\newcommand{\ntrace}{\bar{\trace}\,}

\newcommand{\Entrace}{\mathbb E\,\ntrace}
\newcommand{\dof}{\operatorname{df}}

\usepackage{mdframed}
\usepackage{hyperref}
\usepackage{url}
\usepackage{graphicx}
\usepackage{enumitem}
\usepackage{minitoc}
\def\code#1{\texttt{#1}}

\icmltitlerunning{\code{auto-fpt}: Automating Free Probability Theory Calculations for Machine Learning Theory}

\begin{document}

\doparttoc %
\faketableofcontents %

\twocolumn[
\icmltitle{\code{auto-fpt}: Automating Free Probability Theory Calculations for Machine Learning Theory}

\icmlsetsymbol{equal}{*}

\begin{icmlauthorlist}
\icmlauthor{Arjun Subramonian}{xxx}
\icmlauthor{Elvis Dohmatob}{aaa,bbb,ccc}
\end{icmlauthorlist}

\icmlaffiliation{xxx}{University of California, Los Angeles}
\icmlaffiliation{aaa}{Concordia University}
\icmlaffiliation{bbb}{Meta FAIR}
\icmlaffiliation{ccc}{Mila}

\icmlcorrespondingauthor{Arjun Subramonian}{arjunsub@cs.ucla.edu}

\icmlkeywords{learning theory, high-dimensional analysis, free probability theory}

\vskip 0.3in
]

\printAffiliationsAndNotice{} %

\begin{abstract}
A large part of modern machine learning theory often involves computing the high-dimensional expected trace of a rational expression of large rectangular random matrices. To symbolically compute such quantities using free probability theory, we introduce \code{auto-fpt},
a lightweight Python and SymPy-based tool that can automatically produce a reduced system of fixed-point equations which can be solved for the quantities of interest, and effectively constitutes a theory.
We overview the algorithmic ideas underlying \code{auto-fpt} and its applications to various interesting problems, such as the high-dimensional error of linearized feed-forward neural networks, recovering well-known results. We hope that \code{auto-fpt} streamlines the majority of calculations involved in high-dimensional analysis, while helping the machine learning community reproduce known and uncover new phenomena. Our code can be found at: \url{https://github.com/ArjunSubramonian/auto-fpt}.
\end{abstract}

\section{Introduction}
A common task in machine learning (ML) theory is computing the \emph{high-dimensional} limiting value of the expected normalized trace of a rational expression $R$ of rectangular random matrices, i.e.,
\begin{align}
r=\lim_{\phi,\psi,\ldots} \mathbb{E}\, \ntrace R(A,B,\ldots; Z,W,\ldots;\lambda,\alpha, t,\ldots),
\label{eq:rat-expr}
\end{align}
where $\ntrace$ is the normalized $\trace$ operator, which maps a square matrix to the average of its eigenvalues. Equation \ref{eq:rat-expr} contains the following elements:

\textbf{Deterministic matrices:} $A, B, \ldots$ represent deterministic matrices, e.g., the population covariance matrices for multivariate Gaussian distributions.

\textbf{Random matrices:} $Z, W, \ldots$ are rectangular Gaussian random matrices with IID entries from ${\cal N} (0,c)$, for some constant $c>0$. For example, $Z$ could be the whitened version of the design matrix for a regression problem, while $W$ could be the Glorot-initialized weights in the random features or NTK approximation of a neural network.

\textbf{Scaling constants:} $\phi, \psi, \ldots$ are scalars which might describe some data structural constraints or model design choices which must be respected while taking the limit. These include the aspect ratio $d/n$ (i.e., rate of features to samples) and parametrization $m/n$ (i.e., rate of parameters to samples).

\textbf{Practitioner interventions:} $\lambda, \alpha, t$ are scalars which capture the interventions of ML practitioners. This includes the choice of regularization penalty, and step size and stopping time for optimization algorithms.
This also encompasses the coefficients of Hermite polynomials corresponding to the choice of activation functions in the case of neural networks.

$R$ typically captures a quantity like generalization or population test error \citep{hastie2022surprises,dobriban2018high, Adlam2020DoubleDescent,bach2024high}, adversarial robustness error \citep{pmlr-v206-scetbon23a}, out-of-distribution error \citep{Tripuraneni2021OOD}, or subgroup test error \citep{subramonian2025an}. The sought-for limit $r$ is often considered in a regime where the dimensions of the rectangular matrices diverge at given proportionate rates, e.g., $d/n \to \phi$, $m/n \to \psi$, where $d$ is the input dimension, $n$ is the sample size, and $m$ is the width of the neural network.
To an ML theorist, constants like $\phi$, $\psi$, and $\lambda$ play an analogous role to the constants that appear in the standard model of physics (e.g., mass of electron, charge of electron).
A theorist can analyze $r$ in terms of constants like $\phi$, $\psi$, and $\lambda$ to obtain fundamental insights into ML.

In practice, the limit $r$ can be precisely computed analytically using the framework of operator-valued free probability theory (FPT) \citep{mingo2017free}, which is at the intersection of random matrix theory (RMT) and functional analysis. An alternative approach is via replica symmetry (RS) calculations. Though analytically powerful, RS is not rigorous and its proper use in ML theory requires some training in statistical physics. In comparison, FPT only requires knowledge of linear algebra and statistics.

\textbf{Main contributions.} We introduce and document a lightweight Python-based tool called \code{auto-fpt} for automatically computing $r$ symbolically, as a function of all the relevant constants. We further show how to use \code{auto-fpt} to compute the high-dimensional: (1) Stieltjes transform of the Marchenko-Pastur law, (2) generalization error of classical ridge regression, (3) sum of two sample covariance matrices (subordination), and (4) training error of a random features model.
Our tool takes as input a linear pencil for $R$ and outputs a reduced system of fixed-point equations that can be solved to obtain $r$.
We introduce various techniques to make the computation of these equations more tractable, especially for large pencils, e.g., matrix scalarization, sparse block matrix inversion, and duplicate equation identification and pruning. \code{auto-fpt} currently assumes that the random matrices are Gaussian with IID entries and the deterministic matrices commute.

A primary goal of \code{auto-fpt} is to make the machinery of FPT more accessible to ML theorists and practitioners, so that they can focus on other important aspects of the theoretical analysis of ML models, rather than tedious case-by-case algebraic manipulations.
To the best of our knowledge, there do not exist any open-access tools that automatically symbolically compute the high-dimensional limiting value of the expected normalized trace of rational expressions from their pencils (see Section \ref{sec:rw}).
The only dependencies required by \code{auto-fpt} are SymPy \citep{sympy} and NumPy \citep{harris2020array}, both of which are well-established open-source Python libraries. 
We hope that the ML community leverages \code{auto-fpt} to reproduce known and uncover new phenomena in ML that contribute to our foundational understanding of phenomena like generalization, neural scaling laws, feature learning, memorization, grokking, model collapse, and algorithmic fairness, in solvable regimes which are relevant to neural networks and large language models. %
This is especially important in the context of emerging practices (e.g., training on synthetic data, using attention, adapting to human feedback).

\section{Related Work}
\label{sec:rw}

\textbf{High-dimensional analysis of ML with FPT.}
A suite of works have used operator-valued FPT to analyze the expected behavior of ML models in appropriate asymptotic scaling limits, e.g., the rate of features $d$ to samples $n$ converges to a finite values as $d$ and $n$ respectively scale towards infinity. For example, 
\citet{Adlam2020DoubleDescent} theoretically analyze the double descent phenomenon \citep{spigler2019jamming,belkin2019reconciling} in ridge regression with random projections in a proportionate scaling limit. 
\citet{adlam2020neural} precisely characterize the behavior of wide neural networks (e.g., triple descent). \citet{Tripuraneni2021OOD} present a high-dimensional analysis of how covariate shifts affect the generalization of random feature regression in different parameterization regimes. \citet{lee2023demystifying} analyze the disagreement of random feature models in high dimensions. 
\citet{dohmatob2025strong,dohmatob2021fundamentaltradeoffsmemorizationrobustness} precisely characterize the model collapse phenomenon in the random features regime, showing that even a small fraction of synthetic training data can yield a performance degradation. 
\citet{subramonian2025an} study how random feature models amplify data biases in different parameterization regimes. \code{auto-fpt} can be used to automatically recover the FPT components of the proofs in the aforementioned works, and expand upon their results.

\textbf{Automating FPT computations.} NCAlgebra \citep{ncalgebra} offers functionality to compute minimal linear pencils (specifically, descriptor realizations) for rational expressions. However, to our knowledge, \code{auto-fpt} is the first open-source and open-access tool to automatically symbolically compute the high-dimensional limiting value of the expected trace of rational expressions from their pencils. Our \code{auto-fpt} tool aims to fill this gap using FPT, in a manner that supports ML theory research. We note that the primary contribution of \code{auto-fpt} is {\em not} pencil computation.

\section{A Motivating Example: Marchenko-Pastur}
\label{sec:mp}
As an example of the functionality of \code{auto-fpt}, we briefly show how to use the tool to derive the Stieltjes transform of the MP law  \citep{MP1967} in just a few lines of Python code. In short, the limiting Stieltjes transform $s(\lambda)$ of the spectral distribution of the empirical covariance matrix $X^\top X/n$ can be written as 
\begin{align}
s(\lambda) &= \lim_{\substack{n,d \to \infty\\d/n\to\phi}}  \Entrace (X^\top X/n + \lambda I_d)^{-1} \\
&= \lim_{\substack{n,d \to \infty\\d/n\to\phi}} \Entrace (Z^\top Z + I_d)^{-1}/\lambda.
\end{align}
 Here, $X$ is an $n \times d$ random design matrix with IID entries from $\mathcal N(0,1)$, $Z=X/\sqrt{n \lambda}$, and $\lambda$ is a scalar.
 We use \code{auto-fpt} to generate fixed-point equations for the trace $s(\lambda)$ of the resolvent matrix $(Z^\top Z + I_d)^{-1}$.

\begin{lstlisting}[language=python,numbers=none,basicstyle=\ttfamily\scriptsize]
from sympy import Symbol, MatrixSymbol, Identity
import numpy as np
from fpt import calc

# Form the design matrix.
n = Symbol("n", integer=True, positive=True)
d = Symbol("d", integer=True, positive=True)
Z = MatrixSymbol("Z", n, d)
phi = Symbol(r"\phi", positive=True)

# Form the expression of the resolvent.
MP_expr = (Z.T * Z + Identity(d)).inv()

# Compute a minimal linear pencil using NCAlgebra.
Q, (u, v) = compute_minimal_pencil(MP_expr)

# Get the index of the one-hot entry in u, v.
i = np.flatnonzero(u)[0]
j = np.flatnonzero(v)[0]

# Get free probability equations defining the limiting
# value of the trace of the resolvent. The
# normalize="full" option allows us to apply
# the variance 1/(n*lambda) of the entries of the random
# matrix Z.
calc(Q, random_matrices="Z", row_idx=i, col_idx=j, 
                subs={d: n * phi}, normalize="full")
\end{lstlisting}
Under the hood, NCAlgebra can be used to compute a minimal pencil for $(Z^\top Z + I_d)^{-1}$. This code snippet yields:
\begin{gather*}
    Q = \left[\begin{matrix}I & - Z\\Z^\top  & I\end{matrix}\right], \quad
    u = [0, 1],\quad v = [0, 1],
\end{gather*}
and the fixed-point equations directly below, where $G$ is a $2 \times 2$ matrix which contains the (limiting values) of the expected normalized traces of the blocks of $Q^{-1}$. In particular, $s(\lambda) = G_{1,1}$ (with zero-based indexing), where:
\begin{align}
{G}_{1,1} = \frac{\lambda}{\lambda + {G}_{0,0}},\quad
{G}_{0,0} = \frac{\lambda}{\phi {G}_{1,1} + \lambda}.
\label{eq:mp-eqs}
\end{align}

With minimal substitution and manipulation, we correctly arrive at the following well-known fixed-point equation for the Stieltjes transform of the MP law:
\begin{eqnarray}
\frac{1}{s(\lambda)} = \lambda + \frac{1}{1 + \phi s(\lambda)}.
\end{eqnarray}

We analyze the anisotropic MP law in Appendix \ref{sec:anisotropic-MP-law}.

\section{Preliminaries for Ridge Regression}

For simplicity and as a specific running example of $r$ with which to explain \code{auto-fpt}, we focus on the high-dimensional generalization error of ridge regression. 

\subsection{Setup}
\label{sec:models-metrics}
\paragraph{Data Distribution.}
We consider the following multivariate Gaussian setup \cite{hastie2022surprises,richards2021asymptotics,dobriban2018high,bach2024high}:
\begin{align}
    &\textbf{(Features) }x \sim \mathcal N(0, \Sigma),\\
    &\textbf{(Ground-truth weights) }w^* \sim \mathcal N(0, \Theta/d), \\
    &\textbf{(Labels) }y \mid x \sim \mathcal N(f^\star (x),\sigma^2),\, f^\star(x) := x^\top w^*.
\end{align}
The $d \times d$ symmetric positive-definite matrix $\Sigma$ is the population covariance matrix. The $d$-dimensional vector $w^*$ is the ground-truth weight vector, and it is sampled from a zero-mean Gaussian distribution with covariance matrix $\Theta / d$. Finally, $\sigma^2$ corresponds to the label-noise level.

\paragraph{Models and Metrics.}
We consider the solvable setting of classical ridge regression in the ambient input space. In addition to its analytical tractability, ridge regression can be viewed as the asymptotic limit of many learning problems \citep{dobriban2018high, richards2021asymptotics, hastie2022surprises}. We take the function class $\mathcal F \subseteq \{\mathbb R^d \to \mathbb R\}$ of linear ridge regression models.
For $w \in \mathbb R^d$, the model $f$ is defined by $f(x) = x^\top w,\text{ for all }x \in \mathbb R^d$, and is learned with $\ell_2$-regularization. We further define the generalization error of any model $f$ as $R (f) = \mathbb E\,[(f(x) - f^\star(x))^2].$

\textbf{Learning.} A learner is given $n$ IID samples $\mathcal D = \{(x_1,y_1),\ldots,(x_n,y_n)\} = (X \in \mathbb{R}^{n \times d}, Y \in \mathbb{R}^n)$ of data from the above distribution and it learns a model for predicting the label $y$ from the feature vector $x$. Thus, $X$ is the design matrix with $i$th row $x_i$, and $y$ the response vector with $i$th component $y_i$. We learn the classical ridge regression model $\widehat f$ using empirical risk minimization and $\ell_2$-regularization with penalty $\lambda$. The parameter vector $\widehat w \in \mathbb R^d$ of the linear model $\widehat f:x \mapsto x^\top \widehat w$ is given by:
\begin{align}
\widehat w &= \arg\min_{w \in \mathbb R^d} n^{-1} \|X w - Y\|^2_2 + \lambda \|w\|_2^2 \\
&= (M + \lambda I_d)^{-1} X^\top Y / n,
\label{eq:classical-ridge}
\end{align}
where $M = X^\top X / n$ is the empirical covariance matrix.
This is the ridge estimator; the unregularized limit  $\lambda \to 0^+$ corresponds to ordinary least-squares (OLS). We can decompose the generalization error of $\widehat f$ into bias and variance terms as $R(\widehat f) = B(\widehat f) + V(\widehat f)$ where:
\begin{align}
B(\widehat f) &= \mathbb E\,\|(M + \lambda I_d)^{-1} M w^* -w^*\|_{\Sigma}^2,\\
V(\widehat f) &=  \mathbb E\,\|(M + \lambda I_d)^{-1} X^\top E / n\|_{\Sigma}^2.
\end{align}
Here, $E := Y - X w^*$ is the noise vector, a random $n$-dimensional vector with IID components from $N(0,\sigma^2)$, and independent of $X$ and $w^*$. Further expanding the above expressions for $B(\widehat f)$ and $V(\widehat f)$,
we obtain:
\begin{align}
    B(\widehat f) &= \mathbb E\,\|(M + \lambda I_d)^{-1} M w^* \\
    &-(M + \lambda I_d)^{-1} (M + \lambda I_d) w^*\|_{\Sigma}^2\nonumber \\
    &= \lambda^2 \mathbb E\,\|(M + \lambda I_d)^{-1} w^*\|_{\Sigma}^2\nonumber\\
    &= \lambda^2 \mathbb E\, \ntrace (M + \lambda I_d)^{-1} \Theta (M + \lambda I_d)^{-1} \Sigma, \\
    V(\widehat f)
    &= \frac{\sigma^2}{n} \mathbb E\, \trace (M + \lambda I_d)^{-1} M (M + \lambda I_d)^{-1} \Sigma.
\end{align}

\paragraph{Proportionate Scaling Limit.}
The analysis will be carried out in a proportionate scaling limit (standard in random matrix theory) which enables us to derive deterministic formulas for the expected test risk of models:
\begin{eqnarray}
d,n \to \infty,\quad d/n \to \phi.
\end{eqnarray}
The constant $\phi \in (0,\infty)$ then captures the rate of features to samples. %

\subsection{Free Probability Theory Objective}
Ultimately, our objective is to compute $\lim_{\phi} R(\widehat f) = \lim_{\phi} B(\widehat f) + \lim_{\phi} V(\widehat f)$, where $\lim_\phi$ is shorthand for $\lim_{\substack{n,d \to \infty\\d/n\to\phi}}$.
We define $Z \in \mathbb R^{n \times d}$ to be a random matrix with IID entries sampled from ${\cal N} (0, 1 / \sqrt{n \lambda})$. We define $K := \Sigma^{1/2} Z^\top Z \Sigma^{1/2} + I_d$. Then, by re-expressing the bias and variance terms in the form of Eqn. \ref{eq:rat-expr}, we obtain the following limits of the expected trace of rational functions:
\begin{align}
    \lim_{\phi} B(\widehat f) &= r_B = \lim_{\phi} \mathbb E\, \ntrace K^{-1} \Theta K^{-1} \Sigma, \\
    \lim_{\phi} V(\widehat f) &= \sigma^2 \lambda \phi^{-1} r_V, \text{ where } \\
   r_V &= \lim_{\phi} \mathbb E\,\ntrace K^{-1} \Sigma^{1/2} Z^\top Z \Sigma^{1/2} K^{-1} \Sigma.
\end{align}
These computations will be carried out in Section \ref{sec:case-study-ridge}.

\section{Case Study: Classical Ridge Regression}
\label{sec:case-study-ridge}
In this section, we use our tool to obtain a full analytical picture for the generalization error for classical ridge regression. Particularly, we shall recover the results of \citet{dobriban2018high,richards2021asymptotics,hastie2022surprises}.

\subsection{Computing Minimal Linear Pencil}
\label{sec:compute-pencil}
The first step to applying FPT to compute $r_B$ and $r_V$ is constructing minimal linear pencils for their corresponding rational matrix functions: $R_B = K^{-1} \Theta K^{-1} \Sigma$, $R_V = K^{-1} \Sigma^{1/2} Z^\top Z \Sigma^{1/2} K^{-1} \Sigma.$

\begin{definition}[Linear Pencil \citep{HELTON2006105,HELTON20181,volvcivc2024linear}]
For a $p \times p$ block matrix and vectors $u, v \in \mathbb R^p$, $(u, Q, v)$ is a linear pencil for a rational function $R(A, B, \ldots; Z, W, \ldots)$ if each entry of $Q$ is an affine function of the input matrices and $R = u^\top Q^{-1} v$. %
\end{definition}

With a na{\"i}ve approach to constructing pencils, the resultant pencil may be overly large, yielding an intractable number of fixed-point equations. Hence, we seek minimal pencils w.r.t to the size $p$, which always exist \citep{HELTON2006105,HELTON20181}.
NCAlgebra yields the pencil in Appendix \ref{sec:ridge-reg-pencils-bias}. One can verify that $R_B = Q^{-1}_{3, 8}$. Additional simplifications can be made programmatically, such as reverting $\Sigma_{sqrt}$ to $\Sigma^{1/2}$. We recall that the mechanical advantage of $Q$ over $R_B$ is that $Q$ is linear in the random matrix $Z$. We follow the same process to get a minimal linear pencil for $r_V$ (displayed in Appendix \ref{sec:ridge-reg-pencils-var}).

\subsection{Constructing Fixed-Point Equations via Freeness}

Now, to compute the fixed-point equations that solve for both $r_B$ and $r_V$, we can simply run:
\begin{lstlisting}[language=python,numbers=none,basicstyle=\ttfamily\scriptsize]
# Get the index of the one-hot entry in u, v.
bias_i = np.flatnonzero(bias_u)[0]
bias_j = np.flatnonzero(bias_v)[0]

# Compute fixed-point equations for bias term using FPT.
eqns = calc(bias_Q, random_matrices="Z", row_idx=bias_i,
    col_idx=bias_j, normalize="full", subs={d: n * phi})

# Get the index of the one-hot entry in u, v.
var_i = np.flatnonzero(var_u)[0]
var_j = np.flatnonzero(var_v)[0]

# Compute the fixed-point equations for the
# variance term using FPT.
eqns = calc(var_Q, random_matrices="Z", row_idx=var_i, 
    col_idx=var_j, normalize="full", subs={d: n * phi})
\end{lstlisting}
This code snippet yields the system of fixed-point equations displayed in Figure \ref{fig:bias-var-eqs}. Here, \code{(bias\_u, bias\_Q, bias\_v)} and \code{(var\_u, var\_Q, var\_v)} are the pencils in Appendix \ref{sec:ridge-reg-pencils} 
for the bias and variance terms of the ridge regression generalization error, respectively. The sought-for normalized trace corresponds to entry \code{(row\_idx, col\_idx)} (with zero-based indexing) of the inverse of $Q$. In addition, we need to specify that $Z$ is the random matrix appearing in $Q$ and requires ``full'' normalization (i.e., it has a variance of $\frac{1}{n \lambda}$). Notably, \code{calc} returns the fixed-point equations (in SymPy format), which can be directly displayed in LaTeX (using SymPy's \code{latex} function) or used programmatically (e.g., to validate that the theoretical result correctly predicts experimental findings).

The setup of \code{auto-fpt} necessitates that $u, v$ are one-hot.
However, the pencils computed by NCAlgebra may not have one-hot $u, v$. Furthermore, $u$ or $v$ may contain negative entries. However, this behavior is still compatible with \code{auto-fpt}. Specifically, for any pencil $(u, Q, v)$, we can express $u^\top Q^{-1} v$ as a linear combination:
\begin{align}
u^\top Q^{-1} v=\sum_{i,j} u_i v_j \delta_i^\top Q^{-1} \delta_j=\sum_{i,j} u_i v_j Q^{-1}_{i, j},
\end{align}
where $\delta_i$ is the one-hot encoding of $i$. In effect, a pencil for a rational expression $R$ can be easily decomposed into a pencil for each term in $R$. Furthermore, in the case of negative entries, \code{auto-fpt} can be used to compute $-r$ and the equation output for $-r$ can simply be negated.

While the program technically does not require a minimal pencil, we recommend that it is minimal for computational tractability, as SymPy can stall when computing the inverse of large symbolic matrices. The normalization factor can be adjusted per random matrix by appropriately modifying the \code{variances} dictionary in the \code{calc} function in \code{fpt.py}; a verbosity of 1 will output the variance of the entries of each random matrix for confirmation. This program does require that the random matrices are Gaussian with IID entries and assumes that the deterministic matrices commute.

The equations in Figure \ref{fig:bias-var-eqs} (in the appendix) can be further reduced with minimal substitution:
\begin{gather*}
    r_B = \lambda \ntrace \Sigma (\lambda \Theta + \rho_B \Sigma) (e \Sigma + \lambda I_d)^{-2}, \\
    r_V = \lambda (e - \rho_V) \ntrace \Sigma^2 (e \Sigma + \lambda I_d)^{-2},
\end{gather*}
where the constants $(e, \rho_B, \rho_V)$ are the positive solutions to the following system of fixed-point equations:
\begin{gather*}
    \label{eq:e}
    e = \frac{1}{1 + \phi \ntrace \Sigma (e \Sigma + \lambda I_d)^{-1}}, \\
    \rho_B = \phi e^2 \ntrace \Sigma (\lambda \Theta + \rho_B \Sigma) (e \Sigma + \lambda I_d)^{-2}, \\
    \rho_V = \phi e^2 \ntrace \Sigma (\lambda I_d + \rho_V \Sigma) (e \Sigma + \lambda I_d)^{-2}.
\end{gather*}

To be precise, we notice in $r_B$ that $e = G_{1, 1} = G_{6, 6}$ and $\rho_B = G_{1, 6}$. Similarly, in $r_V$, we notice that $e = G_{1, 1} = G_{5, 5}$ and $\rho_V = G_{1, 5}$.
These implicit equations are not analytically solvable for general matrices.

By manipulating the above equations, we can recover the formulae reported by \citet{bach2024high} (see Proposition 3 therein).
Define $\dof_m (t) := \trace \Sigma^m (\Sigma + t I_d)^{-m}$ and set $\kappa := \lambda / e$. With a bit of algebra, one can now derive from Eqn. \ref{eq:e}:
\begin{align}
    \label{eq:fund}
    \kappa - \lambda = \kappa \dof_1 (\kappa) / n.
\end{align}
Next, we expand $r_B, \rho_B$ and $r_V, \rho_V$ and perform substitution to obtain the equations in Appendix \ref{sec:ridge-reg-eqs-val}. Putting $r_B$ and $r_V$ together, we recover the classical bias-variance decomposition for $\lim_\phi R(\widehat f)$:
\begin{align}
     R(\widehat f) \simeq \frac{\kappa^2 \ntrace \Theta \Sigma (\Sigma + \kappa I_d)^{-2}}{1 - \dof_2 (\kappa) / n} + \frac{\sigma^2 \dof_2 (\kappa) / n}{1 - \dof_2 (\kappa) / n}.
\end{align}

\textbf{Overview of Algorithm.}
We now overview the algorithm and optimizations that \code{auto-fpt} uses to compute the fixed-point equations from $Q, (i, j)$, with the running example of the MP law.
At a high level, our implementation decomposes the input pencil into deterministic and random block matrices, symmetrizes the matrices, and $R$-transforms the random block matrix to obtain a system of equations that can be solved for $r$.

\textbf{Step 1 (Pencil Splitting):} The block matrix $Q$ is split into $Q = F - Q_X$, where $F$ contains the deterministic blocks of $Q$ while $Q_X$ contains the random blocks. This step of the calculations is handled by the a function \code{free\_proba\_utils.decompose\_Q}. For example, if $Q=\left[\begin{matrix}I_n & - Z\\Z^\top  & I_d\end{matrix}\right]$ is the linear pencil matrix we constructed for the MP law in Section \ref{sec:mp}, then
    $$
F=\left[\begin{matrix}I_n & 0\\0  & I_d\end{matrix}\right],\,Q_X=\left[\begin{matrix}0 & Z\\-Z^\top  & 0\end{matrix}\right].
    $$

\textbf{Step 2 (Scalarization):} The block matrices $Q, F, Q_X$ are converted to %
    ordinary matrices $q, f, q_X$ with entries taking values over an abstract commutative ring, respectively, where each unique matrix symbol in $Q, F, Q_X$ is mapped to a different scalar. This step minimizes the runtime of the rest of the program, as SymPy is more optimized for symbolic computation with scalars. Henceforth, we use lowercase to refer to the scalarized version of block matrices. This step is handled by \code{free\_proba\_utils.scalarize\_block\_matrix}.
For the running example, this gives
\begin{align*}
 &f=\left[\begin{matrix}1 & 0\\0  & 1\end{matrix}\right]=I_2,\quad   q = \left[\begin{matrix}1 & -x_1\\x_1  & 1\end{matrix}\right], \\
 &q_x = f-q = \left[\begin{matrix}0 & x_1\\-x_1  & 0\end{matrix}\right],\\
 &\text{ with the identification }Z,Z^\top \to x_1.
\end{align*}
    
\textbf{Step 3 (Symmetrization):} The matrices $Q, q, F, f, Q_X, q_X$ are symmetrized, which is necessary to compute their $R$-transform. We symmetrize a matrix $A$ as:
      $  A' = \begin{bmatrix}
            0 & A^\top \\
            A & 0
        \end{bmatrix}.
        $
    This step is handled by \code{free\_proba\_utils.symmetrize\_block\_matrix}. For the running example, this gives
\begin{align*}
&F'=\left[\begin{matrix}0 & 0 & I_n & 0 \\ 0 & 0 & 0 & I_d \\ I_n & 0 & 0 & 0\\0  & I_d & 0 & 0\end{matrix}\right], \,
 q_x' = \left[\begin{matrix}0 & 0 & 0 & -x_1 \\ 0 & 0 & x_1 & 0 \\ 0 & x_1 & 0 & 0\\-x_1  & 0 & 0 & 0\end{matrix}\right].
\end{align*}
    
\textbf{Step 4 ($R$-transform Computation):}
    Using operator-valued FPT, we obain that in the proportionate scaling limit:
    \begin{eqnarray}
    G = (I_p \otimes \ntrace)[(F' -  \mathcal R_{Q_X'} (G)^\circ)^{-1}],
    \end{eqnarray}
    where $p$ is the number of blocks in $Q'$. Here, $\mathcal R_{Q'_X}$ is the $R$-transform of $Q'_X$ which maps $\mathbb R^{p \times p}$ to itself, and is defined further below. The operator $I_p \otimes \ntrace$ extracts the normalized traces of the blocks of a block matrix, with the convention that the trace of a rectangular block is zero, while the operation $B \mapsto B^\circ$ embeds a $p \times p$ matrix $B$ into an element of the algebra of block matrices having the same size and block shapes as $Q'$ (and therefore $Q_X'$), and is defined in Appendix \ref{sec:algebra}. When $Q'_X$ is a Gaussian block matrix, we have
    \begin{equation}
        \mathcal R_{Q'_X}(G)_{ij} = \sum_{k,\ell}\sigma(i,k;\ell ,j)\alpha_k g_{k\ell},
        \label{eq:Rtransform}
    \end{equation}
    where $\sigma(i,k;\ell,j)$ is the covariance between the entries of block $(i,k)$ and block $(\ell,j)$ of $Q'_X$, while $\alpha_k$ is the dimension of the block $(k,\ell)$ of $G$ \citep{Rashidi2008MIMO}. This step is handled by \code{free\_proba\_utils.R\_transform}.
    
    For the running example, this gives
    \begin{align*}
        &\mathcal R_{Q_X'} (G)^\circ \\
        &=\left[\begin{matrix}0 & 0 & -\frac{d G_{1, 1}}{\lambda n} I_n & 0 \\ 0 & 0 & 0 & -\frac{G_{0, 0}}{\lambda} I_d \\ -\frac{d G_{1, 1}}{\lambda n} I_n & 0 & 0 & 0\\0  & -\frac{G_{0, 0}}{\lambda} I_d & 0 & 0\end{matrix}\right].
    \end{align*}

    To greatly increase the efficiency of the $R$-transform computation and the inversion in the next step, we leverage the sparse structure of $(Q')^{-1}$ to infer the structure of $G$ and avoid computation for $0$-entries. 0-entries in $Q^{-1}$ correspond to 0-entries in $G$ since $G = (I_p \otimes \ntrace)(Q')^{-1}$. By construction, $(Q')^{-1} = \begin{bmatrix}
            0 & Q^{-1} \\ (Q^{-1})^\top & 0
    \end{bmatrix}$; hence, we just need to find the structure of $Q^{-1}$. We do so by proxy by efficiently computing $q^{-1}$. We split $q$ into a 2-by-2 block matrix $\begin{bmatrix} A & B \\ C & D \end{bmatrix}$ (with equal-sized blocks) and apply the formula for the inverse of such a block matrix. We take advantage of the fact that in many settings, the pencil $Q$ has the majority of its entries along or near its diagonal; hence, $A$ and $D$ are often sparse and SymPy's \code{inv\_quick} function can be applied to these blocks to speed-up computation. In the running example, there are no 0-entries.

\textbf{Step 5 (Difference Inversion):} We now compute $(f - r)^{-1}$, where $r$ is the scalarization of the desymmetrized version of $\mathcal R_{Q_X'} (G)^\circ$. We use the same technique from the previous step to increase the efficiency of the inversion of $f - r$. This step is handled by \code{free\_proba\_utils.inv\_heuristic}. For the running example, this gives
    \begin{align*}
        &(f - r)^{-1} = \left[\begin{matrix} \frac{1}{\frac{d}{\lambda n} G_{1, 1} + 1} & 0 \\0  & \frac{1}{\frac{1}{\lambda} G_{0, 0} + 1}\end{matrix}\right].
    \end{align*}

\textbf{Step 6 (Fixed-Point Equation Construction):} The relevant fixed-point equations are recursively extracted from $g = (f - r)^{-1}$. 
    In particular, the first equation in our system (corresponding to the quantity of interest $r$) is $g_{i, j}$. Then, for each equation $g_{s, t} = f(g)$ in our system, if $f(g)$ contains entries in $g$ for which our system does not contain an equation, we extract the relevant equation from $g = (f - r)^{-1}$. We repeat this procedure until the system of equations is closed.  Then, the equations are simplified, e.g., $d / n$ is replaced with $\phi$. These simplifications can be manually specified in \code{fpt.py} using the \code{subs} dictionary. This step is handled by \code{fpt.construct\_fixed\_point\_equations}. \\

    In practice, the resultant system has redundant equations, as $g$ contains many equivalent entries (e.g., due to the cyclic permutation invariance of trace of matrix products). To prune such equations, we identify equivalence classes of entries in $g$ by identifying equivalent entries of $Q^{-1}$ (from our computation in Step 4). Then, for each equivalence class, we: (1) replace all occurrences of its members in our system with a single representative of the class, and (2) remove all the equations corresponding to members of the class except for the equation for the representative. This procedure greatly reduces the size of the system. In the running example, $Q^{-1}_{0, 0} = Q^{-1}_{1, 1}$. This gives
    \begin{align}
        G_{0, 0} = \frac{\lambda}{\lambda + \phi G_{1, 1}}, \quad G_{1, 1} = \frac{\lambda}{\lambda + G_{0, 0}}.
    \end{align}

\textbf{Step 7 (Matricization):} Scalars in the fixed-point equations corresponding to deterministic matrices (per the map in Step 2) are reverted. In particular, $G$ is recovered from $g$ as $G = (I_p \otimes \ntrace) g$. This step is handled by \code{fpt.matricize\_expr}.
    For the case of our running example, this produces Equation \ref{eq:mp-eqs}.

We note that the runtime of \code{auto-fpt} does not scale with $n, d$, as these dimensions are symbolic. Furthermore, the complexity of inverting the block matrices is dominated by the need for symbolic (rather than numerical) computation and the size of the matrices (i.e., the number of blocks). 

\section{Additional Examples}
We include complete example scripts for using \code{auto-fpt}, including for computing the MP law and generalization error of ridge regression, in the \code{examples} folder of our repository. We provide additional examples below. These computations are tedious with pen and paper, thereby highlighting the benefit of our tool.

\subsection{Subordination: Sum of Independent Empirical Covariance Matrices}
We now turn to the computation of $\Entrace R$, where $R=(M+\lambda I_d)^{-1}$, $M:=M_1+M_2$, and $M_j=X_j^\top X_j/n$, where $X_1$ and $X_2$ are independent, with $X_j$ being an $n_j \times d$ random matrix having IID rows from $\mathcal N(0,\Sigma_j)$. Such calculations come up in the analysis of the model collapse phenomenon \cite{dohmatob2025strong}, where a model is trained on a mixture of synthetic and real data. We work in the proportionate scaling limit:
\[
    d,n_1,n_2 \to \infty,\quad n_1/n \to p_1,\,n_2/n \to p_2,\quad d/n \to \phi,
\]
where $n:=n_1+n_2$. Here $p_1 \in (0,1)$ and $\phi \in (0,\infty)$ are fixed constants and $p_2 = 1-p_1$. \code{auto-fpt} produces the following output (see Appendix \ref{sec:subordination}):
\begin{gather*}
    {G}_{3,3} = \bar{tr}{\left(\lambda \left(\lambda I + p_{1} S_{1}^{2} {G}_{1,1} + p_{2} S_{2}^{2} {G}_{5,5}\right)^{-1} \right)},\\
{G}_{2,0} = \bar{tr}{\left(- \lambda S_{1}^{2} \left(\lambda I + p_{1} S_{1}^{2} {G}_{1,1} + p_{2} S_{2}^{2} {G}_{5,5}\right)^{-1} \right)},\\
{G}_{6,4} = \bar{tr}{\left(- \lambda S_{2}^{2} \left(\lambda I + p_{1} S_{1}^{2} {G}_{1,1} + p_{2} S_{2}^{2} {G}_{5,5}\right)^{-1} \right)},\\
{G}_{1,1} = - \frac{\lambda}{- \lambda + \phi {G}_{2,0}},\quad {G}_{5,5} = - \frac{\lambda}{- \lambda + \phi {G}_{6,4}},
\end{gather*}
where we recall that $S_1^2=\Sigma_1$ and $S_2^2=\Sigma_2$. Observe that, setting $e_1:=G_{1,1}$ and $e_2:=G_{5,5}$ and eliminating $G_{2,0}$ and $G_{6,4}$, we get:
$\ntrace R \simeq G_{3,3}/\lambda = \ntrace K^{-1}$,
where $K:=p_1e_1\Sigma_1 + p_2 e_2 \Sigma_2 + \lambda I_d$, and $e_1,e_2>0$ uniquely solve the system of equations
\begin{align}
    e_1 &= \frac{1}{1+\phi\ntrace\Sigma_1 K^{-1}},\quad
    e_2 = \frac{1}{1+\phi\ntrace\Sigma_2 K^{-1}}.
\end{align}
This recovers an alternative statement of a classical result on the Stieltjes transform of the sum of two independent anisotropic Wishart matrices \citep{MP1967,Kargin2015Subordination}. The above result can be easily extended from two to any finite number of groups by defining: $n := n_1+n_2+\ldots$ and $p_j:=\lim n_j/n \in (0,1)$ for all $j$, with $p_1+p_2+\ldots =1$. The $K$ matrix is now given by $K=p_1 e_1 \Sigma_1 + p_2 e_2 \Sigma_2 + \ldots$, where $\Sigma_j$ is the covariance matrix for the $j$th group. The fixed-point equations for the $e_j$'s are given by
\begin{eqnarray*}
    e_j = \frac{1}{1+\phi\ntrace\Sigma_j K^{-1}},\text{ for all }j.
\end{eqnarray*}
This precisely recovers the main result of \citet{couilletMimo2011}, namely Theorem 1 thereof.

\subsection{High-Dimensional Random Features Model}
We now use \code{auto-fpt} to carry out a  computation done in \citet{adlam2020neural}, which studies high-dimensional random features (RF) approximations of a two-layer neural network learned with a linear signal. The random features model offers a simplified analytic handle for understanding neural networks, e.g., effect of model size, double descent.

As before, let $X=Z \in \mathbb R^{d \times n}$ be a random design matrix, with $n$ columns drawn IID from $\mathcal N(0,I_d)$, and $Y_i \in \mathbb R^{1 \times n}$ be the labels. For each training instance $x_i$ (i.e., column $i$ of $X$), the label $y_i$ is generated as:
\begin{align}
    y_i | x_i, \Omega, \omega \sim \omega \sigma_T (\Omega x_i / \sqrt{d}) / \sqrt{d_T} + \epsilon_i,
\end{align}
where $\sigma_T$ is a coordinate-wise activation function, $\epsilon_i \sim {\cal N}(0, \sigma_\epsilon^2)$ is the label noise, and $\Omega \in \mathbb R^{d_T \times d}$ and $\omega \in \mathbb{R}^{1 \times d_T}$ are parameter matrices with IID entries from ${\cal N} (0, 1)$ that are sampled once for all the data.

We have an initial two-layer neural network $N_0 (x) = W_2 \sigma(W_1 x/\sqrt d) / \sqrt{m}$, where $W_1 \in \mathbb R^{m \times d}$ is the  weights of the hidden layer, with IID entries from $\mathcal N(0,1)$, and $W_2 \in \mathbb R^{1 \times m}$ is the weights of the last layer, with IID entries from $\mathcal N(0,\sigma^2_{W_2})$. For our predictor $\hat{y}$, we consider the class of functions learned by random feature kernel ridge regression with the Neural Tangent Kernel (NTK) \citep{adlam2020neural}. We focus on the training risk
$E_{train} = \mathbb E\,[\frac{1}{n} \| Y - \hat{y} (X) \|_F^2]$ of the network in the following proportionate scaling regime:
\begin{align}
    n,d,m \to \infty,\quad d/n \to \phi,\quad d/m \to \psi,
\end{align}
for some constants $\phi,\psi \in (0,\infty)$. Functionally, we are interested in the training risk, as the test risk $E_{test} = \mathbb E\,[(y - \hat{y} (x))^2]$ can be related to $E_{train}$ by $E_{test} = (\lambda \tau_1)^{-2} E_{train} - \sigma_\epsilon^2$ (where $\tau_1$ is defined by the system of equations \ref{eq:t1-t2}). This relation can be derived via additional free probability theory calculations (see appendix of \citet{adlam2020neural}).

We focus on the setting $\sigma^2_{W_2} \to 0$ (as in Section 6.5 of \citet{adlam2020neural}). We shall need the following nonnegative constants:
\begin{eqnarray}
    \beta := \eta-\zeta,\quad \eta := \mathbb E[\sigma(z)^2],\quad \zeta := [\mathbb E\,\sigma'(z)]^2,
\end{eqnarray}
for $z \sim \mathcal N(0,1)$. Observe that for 
a linear activation $\sigma(z) \equiv z$. Importantly,  $\beta$ captures the degree of curvature/nonlinearity in the activation function $\sigma$; $\beta=0$ corresponds to the linear case of random projections considered in \citet{bach2024high,dohmatob2025strong,subramonian2025an}.

In Appendix \ref{sec:rf-model}, we show that the training error of the network in the NTK regime is given by:
\begin{align}
E_{train} = -\gamma^2 (\tau_2' + \sigma_\epsilon^2 \tau_1'),
\end{align}
where the derivatives of $\tau_1, \tau_2$ are taken with respect to $\lambda$. Using \code{auto-fpt}, we obtain the following fixed-point equations relating $\tau_1, \tau_2$:
\begin{align*}
\lambda \tau_2 &= \frac{\lambda {G}_{0,0}}{\beta {G}_{4,4} + \lambda {G}_{3,0} + \lambda},\quad
{G}_{3,0} = \frac{\phi \zeta {G}_{4,4}}{\lambda \phi + \zeta {G}_{2,2} {G}_{4,4}},\\
\lambda \tau_1 &= \frac{\lambda}{\beta {G}_{4,4} + \lambda {G}_{3,0} + \lambda},\\
{G}_{4,4} &= - \frac{\lambda \phi}{- \beta \psi {G}_{2,2} - \lambda \phi + \phi \psi \zeta {G}_{0,3}},\\
{G}_{0,3} &= - \frac{\lambda {G}_{2,2}}{\lambda \phi + \zeta {G}_{2,2} {G}_{4,4}},\quad
{G}_{0,0} = \frac{\lambda \phi}{\lambda \phi + \zeta {G}_{2,2} {G}_{4,4}}.
\end{align*}

In Appendix \ref{sec:rf-model}, we show via variable elimination that this system exactly recovers the coupled polynomial equations for $\tau_1, \tau_2$ in \citet{adlam2020neural}, namely Proposition 1 thereof with $\sigma^2_{W_2} = 0$:
\begin{align}
    0 &= - \eta \phi \tau_{1}^{2} + \eta \phi \tau_{1} \tau_{2} - \lambda \tau_{1}^{2} \tau_{2} \zeta + \phi \tau_{1}^{2} \zeta \nonumber \\
    &\quad\,- 2 \phi \tau_{1} \tau_{2} \zeta + \phi \tau_{2}^{2} \zeta + \tau_{1} \tau_{2} \zeta, \\
    0 &= \lambda \psi \tau_{1}^{2} \tau_{2} \zeta - \phi^{2} \tau_{1} + \phi^{2} \tau_{2} + \phi \tau_{1} \tau_{2} \zeta - \psi \tau_{1} \tau_{2} \zeta.
    \label{eq:t1-t2}
\end{align}

This calculation requires computing a single pencil for $\tau_1, \tau_2$ (see Appendix \ref{sec:rf-model}). While NCAlgebra does not explicitly support computing a single pencil for multiple rational expressions, the modularity of pencil construction can be exploited to do so. In particular, if we would like to compute a single minimal pencil for the rational expressions $R_A, R_B$, we can instead compute a minimal pencil for $R = R_A + R_B$ and may be able to use the non-zero entries of $u$ and $v$ to identify the relevant blocks of $Q^{-1}$. 

\section{Discussion and Conclusion}
We present \code{auto-fpt}, a lightweight Python-based tool for automatically symbolically computing the high-dimensional expected normalized trace of rational functions from their pencils, which is a common procedure in ML theory.
We introduce various techniques to make this computation more tractable, especially for large pencils, e.g., matrix scalarization, sparse block matrix inversion, and duplicate equation identification and pruning.
We show how to use \code{auto-fpt} to recover known theoretical results.
While we do not present new theoretical results, a preliminary version of \code{auto-fpt} was used to guide the derivation of the theories in \citet{subramonian2025an, dohmatob2025strong}.

\code{auto-fpt} offers numerous promising future directions. The runtime and tractability of the tool can be empirically stress-tested with respect to rational expression size, pencil size, and numerical stability. In addition, the code can be profiled to identify performance bottlenecks. Moreover, user studies can be run with ML theorists (both RMT experts and those who are unfamiliar with FPT) to understand the efficacy and limitations of \code{auto-fpt} for their research and deriving novel theoretical results (e.g., analyzing two-layer neural networks beyond initialization).
For example, by default, \code{auto-fpt} does not support powers of traces of rational expressions, and thus cannot currently asymptotically solve the Harish-Chandra-Itzykson-Zuber integral \citep{troiani2022optimal}.
 
With regards to future functionality, \code{auto-fpt} can be developed to work with non-Gaussian random matrices and non-commuting deterministic matrices.
In addition, \code{auto-fpt} can be  developed to provide more explicit support for converting the objectives of classification problems into rational expressions, and support for learning problems where a closed-form solution may not exist.
Furthermore, it can currently require some manual effort to rewrite interesting quantities in ML as the expected trace of a rational expression, which is required to compute a pencil and use \code{auto-fpt}. Therefore, \code{auto-fpt} can be extended to intelligently perform this rewriting.
\code{auto-fpt} can also be extended to: (1) manipulate and automatically solve the system of equations output by the program, and (2) produce phase diagrams for quantities of interest.
In addition, \code{auto-fpt} may require further development to be integrated as a subroutine into proof assistants.

\section*{Broader Impacts Statement}
We hope that \code{auto-fpt} makes the machinery of ML theory more accessible to theorists and practitioners, and can help reproduce known and uncover new phenomena in ML that contribute to our foundational understanding of sociotechnical phenomena like robustness and fairness. \code{auto-fpt} should be used in conjunction with real-world experiments to avoid overclaiming aspects of ML models. \code{auto-fpt} has potential to be used in educational settings to teach RMT.

\section*{Acknowledgments and Disclosures of Funding}

We thank Yizhou Sun and anonymous reviewers for their helpful feedback. AS is supported by an Amazon Science Hub Fellowship.

\bibliography{iclr2025_conference}

\begin{thebibliography}{27}
\providecommand{\natexlab}[1]{#1}
\providecommand{\url}[1]{\texttt{#1}}
\expandafter\ifx\csname urlstyle\endcsname\relax
  \providecommand{\doi}[1]{doi: #1}\else
  \providecommand{\doi}{doi: \begingroup \urlstyle{rm}\Url}\fi

\bibitem[Adlam \& Pennington(2020{\natexlab{a}})Adlam and Pennington]{Adlam2020DoubleDescent}
Adlam, B. and Pennington, J.
\newblock Understanding double descent requires a fine-grained bias-variance decomposition.
\newblock In Larochelle, H., Ranzato, M., Hadsell, R., Balcan, M., and Lin, H. (eds.), \emph{Advances in Neural Information Processing Systems}, volume~33, pp.\  11022--11032. Curran Associates, Inc., 2020{\natexlab{a}}.
\newblock URL \url{https://proceedings.neurips.cc/paper_files/paper/2020/file/7d420e2b2939762031eed0447a9be19f-Paper.pdf}.

\bibitem[Adlam \& Pennington(2020{\natexlab{b}})Adlam and Pennington]{adlam2020neural}
Adlam, B. and Pennington, J.
\newblock The neural tangent kernel in high dimensions: Triple descent and a multi-scale theory of generalization.
\newblock In \emph{International Conference on Machine Learning}, pp.\  74--84. PMLR, 2020{\natexlab{b}}.

\bibitem[Bach(2024)]{bach2024high}
Bach, F.
\newblock High-dimensional analysis of double descent for linear regression with random projections.
\newblock \emph{SIAM Journal on Mathematics of Data Science}, 6\penalty0 (1):\penalty0 26--50, 2024.

\bibitem[Belkin et~al.(2019)Belkin, Hsu, Ma, and Mandal]{belkin2019reconciling}
Belkin, M., Hsu, D., Ma, S., and Mandal, S.
\newblock Reconciling modern machine-learning practice and the classical bias--variance trade-off.
\newblock \emph{Proceedings of the National Academy of Sciences}, 116\penalty0 (32):\penalty0 15849--15854, 2019.

\bibitem[Couillet et~al.(2011)Couillet, Debbah, and Silverstein]{couilletMimo2011}
Couillet, R., Debbah, M., and Silverstein, J.~W.
\newblock A deterministic equivalent for the analysis of correlated mimo multiple access channels.
\newblock \emph{IEEE Trans. Inf. Theor.}, 57, 2011.

\bibitem[Dobriban \& Wager(2018)Dobriban and Wager]{dobriban2018high}
Dobriban, E. and Wager, S.
\newblock High-dimensional asymptotics of prediction: Ridge regression and classification.
\newblock \emph{The Annals of Statistics}, 46\penalty0 (1):\penalty0 247--279, 2018.

\bibitem[Dohmatob(2021)]{dohmatob2021fundamentaltradeoffsmemorizationrobustness}
Dohmatob, E.
\newblock Fundamental tradeoffs between memorization and robustness in random features and neural tangent regimes.
\newblock \emph{arxiv preprint}, 2021.

\bibitem[Dohmatob et~al.(2025)Dohmatob, Feng, Subramonian, and Kempe]{dohmatob2025strong}
Dohmatob, E., Feng, Y., Subramonian, A., and Kempe, J.
\newblock Strong model collapse.
\newblock In \emph{The Thirteenth International Conference on Learning Representations}, 2025.
\newblock URL \url{https://openreview.net/forum?id=et5l9qPUhm}.

\bibitem[Goldt et~al.(2022)Goldt, Loureiro, Reeves, Krzakala, M{\'e}zard, and Zdeborov{\'a}]{goldt2022gaussian}
Goldt, S., Loureiro, B., Reeves, G., Krzakala, F., M{\'e}zard, M., and Zdeborov{\'a}, L.
\newblock The gaussian equivalence of generative models for learning with shallow neural networks.
\newblock In \emph{Mathematical and Scientific Machine Learning}, pp.\  426--471. PMLR, 2022.

\bibitem[Harris et~al.(2020)Harris, Millman, van~der Walt, Gommers, Virtanen, Cournapeau, Wieser, Taylor, Berg, Smith, Kern, Picus, Hoyer, van Kerkwijk, Brett, Haldane, del R{\'{i}}o, Wiebe, Peterson, G{\'{e}}rard-Marchant, Sheppard, Reddy, Weckesser, Abbasi, Gohlke, and Oliphant]{harris2020array}
Harris, C.~R., Millman, K.~J., van~der Walt, S.~J., Gommers, R., Virtanen, P., Cournapeau, D., Wieser, E., Taylor, J., Berg, S., Smith, N.~J., Kern, R., Picus, M., Hoyer, S., van Kerkwijk, M.~H., Brett, M., Haldane, A., del R{\'{i}}o, J.~F., Wiebe, M., Peterson, P., G{\'{e}}rard-Marchant, P., Sheppard, K., Reddy, T., Weckesser, W., Abbasi, H., Gohlke, C., and Oliphant, T.~E.
\newblock Array programming with {NumPy}.
\newblock \emph{Nature}, 585\penalty0 (7825):\penalty0 357--362, September 2020.
\newblock \doi{10.1038/s41586-020-2649-2}.
\newblock URL \url{https://doi.org/10.1038/s41586-020-2649-2}.

\bibitem[Hastie et~al.(2022)Hastie, Montanari, Rosset, and Tibshirani]{hastie2022surprises}
Hastie, T., Montanari, A., Rosset, S., and Tibshirani, R.~J.
\newblock Surprises in high-dimensional ridgeless least squares interpolation.
\newblock \emph{Annals of statistics}, 50\penalty0 (2):\penalty0 949, 2022.

\bibitem[Helton \& de~Oliveira(2017)Helton and de~Oliveira]{ncalgebra}
Helton, J.~W. and de~Oliveira, M.~C., 2017.
\newblock URL \url{https://github.com/NCAlgebra/NC}.

\bibitem[Helton et~al.(2006)Helton, McCullough, and Vinnikov]{HELTON2006105}
Helton, J.~W., McCullough, S.~A., and Vinnikov, V.
\newblock Noncommutative convexity arises from linear matrix inequalities.
\newblock \emph{Journal of Functional Analysis}, 240\penalty0 (1):\penalty0 105--191, 2006.
\newblock ISSN 0022-1236.
\newblock \doi{https://doi.org/10.1016/j.jfa.2006.03.018}.
\newblock URL \url{https://www.sciencedirect.com/science/article/pii/S0022123606001352}.

\bibitem[Helton et~al.(2018)Helton, Mai, and Speicher]{HELTON20181}
Helton, J.~W., Mai, T., and Speicher, R.
\newblock Applications of realizations (aka linearizations) to free probability.
\newblock \emph{Journal of Functional Analysis}, 274\penalty0 (1):\penalty0 1--79, 2018.
\newblock ISSN 0022-1236.
\newblock \doi{https://doi.org/10.1016/j.jfa.2017.10.003}.
\newblock URL \url{https://www.sciencedirect.com/science/article/pii/S0022123617303798}.

\bibitem[Kargin(2015)]{Kargin2015Subordination}
Kargin, V.
\newblock {Subordination for the sum of two random matrices}.
\newblock \emph{The Annals of Probability}, 43\penalty0 (4), 2015.

\bibitem[Lee et~al.(2023)Lee, Moniri, Huang, Dobriban, and Hassani]{lee2023demystifying}
Lee, D., Moniri, B., Huang, X., Dobriban, E., and Hassani, H.
\newblock Demystifying disagreement-on-the-line in high dimensions.
\newblock In \emph{International Conference on Machine Learning}, pp.\  19053--19093. PMLR, 2023.

\bibitem[Mar{\v{c}}enko \& Pastur(1967)Mar{\v{c}}enko and Pastur]{MP1967}
Mar{\v{c}}enko, V.~A. and Pastur, L.~A.
\newblock Distribution of eigenvalues for some sets of random matrices.
\newblock \emph{Mathematics of the USSR-Sbornik}, 1\penalty0 (4):\penalty0 457, apr 1967.

\bibitem[Meurer et~al.(2017)Meurer, Smith, Paprocki, \v{C}ert\'{i}k, Kirpichev, Rocklin, Kumar, Ivanov, Moore, Singh, Rathnayake, Vig, Granger, Muller, Bonazzi, Gupta, Vats, Johansson, Pedregosa, Curry, Terrel, Rou\v{c}ka, Saboo, Fernando, Kulal, Cimrman, and Scopatz]{sympy}
Meurer, A., Smith, C.~P., Paprocki, M., \v{C}ert\'{i}k, O., Kirpichev, S.~B., Rocklin, M., Kumar, A., Ivanov, S., Moore, J.~K., Singh, S., Rathnayake, T., Vig, S., Granger, B.~E., Muller, R.~P., Bonazzi, F., Gupta, H., Vats, S., Johansson, F., Pedregosa, F., Curry, M.~J., Terrel, A.~R., Rou\v{c}ka, v., Saboo, A., Fernando, I., Kulal, S., Cimrman, R., and Scopatz, A.
\newblock Sympy: symbolic computing in python.
\newblock \emph{PeerJ Computer Science}, 3:\penalty0 e103, January 2017.
\newblock ISSN 2376-5992.
\newblock \doi{10.7717/peerj-cs.103}.

\bibitem[Mingo \& Speicher(2017)Mingo and Speicher]{mingo2017free}
Mingo, J.~A. and Speicher, R.
\newblock \emph{Free Probability and Random Matrices}, volume~35 of \emph{Fields Institute Monographs}.
\newblock Springer, 2017.

\bibitem[Rashidi~Far et~al.(2008)Rashidi~Far, Oraby, Bryc, and Speicher]{Rashidi2008MIMO}
Rashidi~Far, R., Oraby, T., Bryc, W., and Speicher, R.
\newblock On slow-fading mimo systems with nonseparable correlation.
\newblock \emph{IEEE Transactions on Information Theory}, 54\penalty0 (2):\penalty0 544--553, 2008.

\bibitem[Richards et~al.(2021)Richards, Mourtada, and Rosasco]{richards2021asymptotics}
Richards, D., Mourtada, J., and Rosasco, L.
\newblock Asymptotics of ridge (less) regression under general source condition.
\newblock In \emph{International Conference on Artificial Intelligence and Statistics}, pp.\  3889--3897. PMLR, 2021.

\bibitem[Scetbon \& Dohmatob(2023)Scetbon and Dohmatob]{pmlr-v206-scetbon23a}
Scetbon, M. and Dohmatob, E.
\newblock Robust linear regression: Gradient-descent, early-stopping, and beyond.
\newblock In Ruiz, F., Dy, J., and van~de Meent, J.-W. (eds.), \emph{Proceedings of The 26th International Conference on Artificial Intelligence and Statistics}, volume 206 of \emph{Proceedings of Machine Learning Research}, pp.\  11583--11607. PMLR, 25--27 Apr 2023.
\newblock URL \url{https://proceedings.mlr.press/v206/scetbon23a.html}.

\bibitem[Spigler et~al.(2019)Spigler, Geiger, d’Ascoli, Sagun, Biroli, and Wyart]{spigler2019jamming}
Spigler, S., Geiger, M., d’Ascoli, S., Sagun, L., Biroli, G., and Wyart, M.
\newblock A jamming transition from under-to over-parametrization affects generalization in deep learning.
\newblock \emph{Journal of Physics A: Mathematical and Theoretical}, 52\penalty0 (47):\penalty0 474001, 2019.

\bibitem[Subramonian et~al.(2025)Subramonian, Bell, Sagun, and Dohmatob]{subramonian2025an}
Subramonian, A., Bell, S., Sagun, L., and Dohmatob, E.
\newblock An effective theory of bias amplification.
\newblock In \emph{The Thirteenth International Conference on Learning Representations}, 2025.
\newblock URL \url{https://openreview.net/forum?id=VoI4d6uhdr}.

\bibitem[Tripuraneni et~al.(2021)Tripuraneni, Adlam, and Pennington]{Tripuraneni2021OOD}
Tripuraneni, N., Adlam, B., and Pennington, J.
\newblock Overparameterization improves robustness to covariate shift in high dimensions.
\newblock In \emph{Advances in Neural Information Processing Systems}, volume~34, pp.\  13883--13897. Curran Associates, Inc., 2021.

\bibitem[Troiani et~al.(2022)Troiani, Erba, Krzakala, Maillard, and Zdeborov{\'a}]{troiani2022optimal}
Troiani, E., Erba, V., Krzakala, F., Maillard, A., and Zdeborov{\'a}, L.
\newblock Optimal denoising of rotationally invariant rectangular matrices.
\newblock In \emph{Mathematical and Scientific Machine Learning}, pp.\  97--112. PMLR, 2022.

\bibitem[Vol{\v{c}}i{\v{c}}(2024)]{volvcivc2024linear}
Vol{\v{c}}i{\v{c}}, J.
\newblock Linear matrix pencils and noncommutative convexity.
\newblock \emph{arXiv preprint arXiv:2407.08450}, 2024.

\end{thebibliography}
\bibliographystyle{icml2025}

\onecolumn

\appendix
\part{Appendix} %

\parttoc %

\section{Ridge Regression Pencils}
\label{sec:ridge-reg-pencils}

\subsection{Bias Term}
\label{sec:ridge-reg-pencils-bias}

\begin{align}
Q &= \left[\begin{matrix}\mathbb{I} & - Z^\top  & 0 & 0 & 0 & 0 & 0 & 0 & 0\\0 & \mathbb{I} & - Z & 0 & 0 & 0 & 0 & 0 & 0\\0 & 0 & \mathbb{I} & - \Sigma_{sqrt} & 0 & 0 & 0 & 0 & 0\\\Sigma_{sqrt} & 0 & 0 & \mathbb{I} & - \Theta & 0 & 0 & 0 & 0\\0 & 0 & 0 & 0 & \mathbb{I} & \Sigma_{sqrt} & 0 & 0 & - \Sigma\\0 & 0 & 0 & 0 & 0 & \mathbb{I} & - Z^\top  & 0 & 0\\0 & 0 & 0 & 0 & 0 & 0 & \mathbb{I} & - Z & 0\\0 & 0 & 0 & 0 & - \Sigma_{sqrt} & 0 & 0 & \mathbb{I} & 0\\0 & 0 & 0 & 0 & 0 & 0 & 0 & 0 & \mathbb{I}\end{matrix}\right], \\
    u &= [0, 0, 0, 1, 0, 0, 0, 0, 0],  \, v = [0, 0, 0, 0, 0, 0, 0, 0, 1],
\end{align}
where $\Sigma_{sqrt} = \Sigma^{1/2}$.

\subsection{Variance Term}
\label{sec:ridge-reg-pencils-var}

\begin{align}
    Q &= \left[\begin{matrix}\mathbb{I} & - Z^\top  & 0 & 0 & 0 & 0 & 0 & 0 & 0\\0 & \mathbb{I} & - Z & 0 & 0 & 0 & 0 & 0 & 0\\0 & 0 & \mathbb{I} & - \Sigma_{sqrt} & 0 & 0 & 0 & 0 & 0\\\Sigma_{sqrt} & 0 & 0 & \mathbb{I} & - \Sigma_{sqrt} & 0 & 0 & 0 & 0\\0 & 0 & 0 & 0 & \mathbb{I} & - Z^\top  & 0 & 0 & 0\\0 & 0 & 0 & 0 & 0 & \mathbb{I} & - Z & 0 & 0\\0 & 0 & 0 & 0 & 0 & 0 & \mathbb{I} & - \Sigma_{sqrt} & 0\\0 & 0 & 0 & 0 & \Sigma_{sqrt} & 0 & 0 & \mathbb{I} & - \Sigma\\0 & 0 & 0 & 0 & 0 & 0 & 0 & 0 & \mathbb{I}\end{matrix}\right], \\
    u &= [0, 0, 0, 1, 0, 0, 0, 0, 0], \, v = [0, 0, 0, 0, 0, 0, 0, 0, 1].
\end{align}

\section{Flags for \code{fpt.py}}

\begin{table}[!ht]
\centering
\begin{tabular}{|c|l|}
    \toprule
   \textbf{Flag}  & \textbf{Description}  \\
   \midrule
   \code{--pencil-file} & path to \code{.pkl} file containing precomputed pencil (required) \\
    \midrule
   \code{--i} & sought-for normalized trace corresponds to entry $(i, j)$ \\
   & (with zero-based indexing) of inverse of $Q$ (required) \\
   \midrule
   \code{--j} & sought-for normalized trace corresponds to entry $(i, j)$ \\
   & (with zero-based indexing) of inverse of $Q$ (required) \\
   \midrule 
    \code{--random-matrix} & specify a random matrix appearing in the pencil \\
    \midrule 
    \code{--normalize} & specify either ``full'' (random matrices have variance of $\frac{1}{n \lambda}$) or \\ 
    & ``sample size'' (random matrices have variance of $\frac{1}{n}$) \\
    \midrule
    \code{--verbose} & specify verbosity level \\
    \bottomrule
\end{tabular}
\caption{Description of flags that can be passed to \code{fpt.py}. These descriptions, usage examples, and additional documentation can be accessed by running \code{python fpt.py --help}.}
\label{tab:fpt-flags}
\end{table}

\code{fpt.py} can also be run via a command-line interface. To compute $r_B$ and $r_V$, we can run:
\begin{lstlisting}[language=bash,numbers=none]
#!/bin/bash
python fpt.py --pencil-file bias.pkl --i 3 --j 8 \
              --random-matrix Z --normalize "full"
python fpt.py --pencil-file var.pkl --i 3 --j 8 \
              --random-matrix Z --normalize "full"
\end{lstlisting}

\code{fpt.py} has the input flags described in Table \ref{tab:fpt-flags}. The program requires a linear pencil matrix $Q$ for $R$ (stored in a pickle file) and the entry $(i, j)$ of $Q^{-1}$ corresponding to the desired expected normalized trace (with zero-based indexing). Optionally, random matrices, a normalization mode, and verbosity level can be specified. For example, for the bias term, we would need to pass \code{--random-matrix=Z --normalize=full}.

\section{Ridge Regression Fixed-Point Equations}

\begin{figure}[ht!]
    \centering
    \begin{subfigure}[t]{\textwidth}
        \centering
        \begin{gather*}
        {G}_{3,8} = \bar{tr}{\left(- \lambda \Sigma \left(\lambda \mathbb{I} + \Sigma_{sqrt}^{2} {G}_{1,1}\right)^{-2} \left(- \lambda \Theta + \Sigma_{sqrt}^{2} {G}_{1,6}\right) \right)},\\
{G}_{2,5} = \bar{tr}{\left(\lambda \Sigma_{sqrt}^{2} \left(\lambda \mathbb{I} + \Sigma_{sqrt}^{2} {G}_{1,1}\right)^{-1} \left(\lambda \mathbb{I} + \Sigma_{sqrt}^{2} {G}_{6,6}\right)^{-1} \left(- \lambda \Theta + \Sigma_{sqrt}^{2} {G}_{1,6}\right) \right)},\\
{G}_{6,6} = - \frac{\lambda}{\phi {G}_{7,5} - \lambda},\\
{G}_{2,0} = \bar{tr}{\left(- \lambda \Sigma_{sqrt}^{2} \left(\lambda \mathbb{I} + \Sigma_{sqrt}^{2} {G}_{1,1}\right)^{-1} \right)},\\
{G}_{1,1} = - \frac{\lambda}{\phi {G}_{2,0} - \lambda},\\
{G}_{1,6} = \frac{\phi \lambda {G}_{2,5}}{\left(\phi {G}_{2,0} - \lambda\right) \left(\phi {G}_{7,5} - \lambda\right)},\\
{G}_{7,5} = \bar{tr}{\left(- \lambda \Sigma_{sqrt}^{2} \left(\lambda \mathbb{I} + \Sigma_{sqrt}^{2} {G}_{6,6}\right)^{-1} \right)}
\end{gather*}
        \caption{Fixed-point equations for $r_B$.}
    \end{subfigure}
    \begin{subfigure}[t]{\textwidth}
        \centering
        \begin{gather*}
           {G}_{3,8} = \bar{tr}{\left(\lambda \Sigma \Sigma_{sqrt}^{2} \left(\lambda \mathbb{I} + \Sigma_{sqrt}^{2} {G}_{1,1}\right)^{-2} \left({G}_{1,1} - {G}_{1,5}\right) \right)},\\
{G}_{6,4} = \bar{tr}{\left(- \lambda \Sigma_{sqrt}^{2} \left(\lambda \mathbb{I} + \Sigma_{sqrt}^{2} {G}_{5,5}\right)^{-1} \right)},\\
{G}_{2,4} = \bar{tr}{\left(\lambda \Sigma_{sqrt}^{2} \left(\lambda \mathbb{I} + \Sigma_{sqrt}^{2} {G}_{1,1}\right)^{-1} \left(\lambda \mathbb{I} + \Sigma_{sqrt}^{2} {G}_{5,5}\right)^{-1} \left(\lambda \mathbb{I} + \Sigma_{sqrt}^{2} {G}_{1,5}\right) \right)},\\
{G}_{1,1} = - \frac{\lambda}{\phi {G}_{2,0} - \lambda},\\
{G}_{1,5} = \frac{\phi \lambda {G}_{2,4}}{\left(\phi {G}_{2,0} - \lambda\right) \left(\phi {G}_{6,4} - \lambda\right)},\\
{G}_{2,0} = \bar{tr}{\left(- \lambda \Sigma_{sqrt}^{2} \left(\lambda \mathbb{I} + \Sigma_{sqrt}^{2} {G}_{1,1}\right)^{-1} \right)},\\
{G}_{5,5} = - \frac{\lambda}{\phi {G}_{6,4} - \lambda}
        \end{gather*}
        \caption{Fixed-point equations for $r_V$.}
    \end{subfigure}%
    \caption{Fixed-point equations for $r_B$ and $r_V$ output by \code{auto-fpt} in terms of the deterministic matrices $\Sigma_{sqrt}$, $\Theta$, and $\Sigma$. $r_B$ and $r_V$ correspond to $G_{3, 8}$.}
    \label{fig:bias-var-eqs}
\end{figure}

\section{Manipulation of Ridge Regression Fixed-Point Equations}
\label{sec:ridge-reg-eqs-val}

We expand $r_B$ and $\rho_B$ and perform substitution to obtain:
\begin{align}
    r_B &= \kappa^2 \ntrace \Theta \Sigma (\Sigma + \kappa I_d)^{-2} + \kappa (\rho_B / e) \dof_2 (\kappa) / d, \\
    \rho_B / e &= \phi \kappa \ntrace \Theta \Sigma (\Sigma + \kappa I_d)^{-2} + (\rho_B / e) \dof_2 (\kappa) / n \\
    &= \frac{\phi \kappa \ntrace \Theta \Sigma (\Sigma + \kappa I_d)^{-2}}{1 - \dof_2 (\kappa) / n},
\end{align}
\begin{align}
    r_B &= \kappa^2 \ntrace \Theta \Sigma (\Sigma + \kappa I_d)^{-2} + \frac{\dof_2 (\kappa) / n}{1 - \dof_2 (\kappa) / n} \cdot \kappa^2 \ntrace \Theta \Sigma (\Sigma + \kappa I_d)^{-2} \\
    &= \frac{\kappa^2 \ntrace \Theta \Sigma (\Sigma + \kappa I_d)^{-2}}{1 - \dof_2 (\kappa) / n}.
\end{align}

Similarly, with $r_V$ and $\rho_V$:
\begin{align}
     r_V / \lambda &= (1 / e - \rho_V / e^2) \dof_2 (\kappa) / d, \\
     \rho_V / e^2 &= \phi (1 / e) \kappa \ntrace \Sigma ( \Sigma + \kappa I_d)^{-2} + (\rho_V / e^2) \dof_2 (\kappa) / n \\
     &= \frac{(1 / e) (\dof_1 (\kappa) - \dof_2 (\kappa)) / n}{1 - \dof_2 (\kappa) / n} \\
     &= \frac{(1 / e - 1) - (1 / e) \dof_2 (\kappa) / n}{1 - \dof_2 (\kappa) / n} \text{ (Eqn. \ref{eq:fund})}, \\
     r_V / \lambda &= 1/e \cdot \dof_2 (\kappa) / d - \frac{(1 / e - 1) - (1 / e) \dof_2 (\kappa) / n}{1 - \dof_2 (\kappa) / n} \cdot \dof_2 (\kappa) / d \\
     &= \frac{\dof_2 (\kappa) / d}{1 - \dof_2 (\kappa) / n}.
\end{align}

\section{An Algebra of Block Matrices}
\label{sec:algebra}

Let $\mathcal A$ be the (von Neumann) non-commutative unital algebra of all block matrices with the same size and block shapes as the $p \times p$ block matrix $Q'$ (and therefore $Q_X'$). Thus, each block $M[i,j]$ of each $M \in \mathcal A$ has same shape $\alpha_i \times \beta_j$ as the corresponding block $Q'[i,j]$ of $Q'$. Let $1_{\mathcal A}$ be the unit element of $\mathcal A$. Note that $\mathcal A$ is fully determined by the $p^2$ pairs of integers $\{(\alpha_i,\beta_j) \mid 1 \le i,j\le p\}$ representing the shapes of the blocks of $Q'$. The operator $I_{p} \otimes \ntrace$ which extracts the normalized traces\footnote{By convention, the trace of a non-square block is zero.} of the blocks of an element $M \in \mathcal A$ is a \emph{tracial state} for this algebra. It maps a block matrix $M \in \mathcal A$ into a $p \times p$ matrix $(I_d \otimes \ntrace)M$.
   
   Any $p \times p$ matrix $B$ can be embedded into an element $B^\circ$ of $\mathcal A$ by defining the $(i,j)$-block of the later as follows:
    \begin{eqnarray}
        B^\circ[i,j] = \begin{cases}
            B_{i,j} I_{\alpha_i},&\mbox{ if }\alpha_i=\beta_j,\\
            0_{\alpha_i\times \beta_j},&\mbox{ otherwise.}
        \end{cases}
    \end{eqnarray}
    It is clear that if $B=I_{p}$, then $B^\circ = 1_{\mathcal A}$.
    
    \textbf{Worked Example.} As an example, observe that if $Q=\left[\begin{matrix}I_n & - Z\\Z^\top  & I_d\end{matrix}\right]$ is the linear pencil we constructed for the MP law in Section \ref{sec:mp}, then by inspecting the shape of the blocks of the symmetrization $Q'$, we get
    $$
    p=4,\, \alpha_1=\alpha_3=\beta_1=\beta_3=n,\, \alpha_2=\alpha_4=\beta_2=\beta_4=d.
    $$
    In this case, any $4 \times 4$ matrix $B$ is mapped to the block matrix $B^\circ \in \mathcal A$ given by
    $$
    B^\circ = \left[\begin{matrix}0 & 0 & B_{1,3}I_n & 0 \\0 & 0 & 0  & B_{2,4}I_d\\
    B_{3,1}I_n & 0 & 0 & 0\\0  & B_{4,2}I_d & 0 & 0\end{matrix}\right].
    $$

\section{Anisotropic MP Law with \code{auto-fpt}}
\label{sec:anisotropic-MP-law}

Consider the case of general covariance matrix $\Sigma$. The design matrix can be written as $X=X_0\Sigma_{sqrt}$, where $\Sigma_{sqrt}=\Sigma^{1/2}$ is the positive square-root of $\Sigma$, i.e., the unique positive-definite matrix such that $\Sigma_{sqrt}^2 = \Sigma$. As before $X_0$ is an $n \times d$ random matrix with IID entries from $\mathcal N(0,1)$. Thus, $X$ has IID rows distributed according to a multivariate Gaussian distribution $\mathcal N(0,\Sigma)$. Our goal is to compute the Stieltjes transform of the empirical covariance matrix $X^\top X/n$, namely the expected value of the trace of the resolvent matrix $R=(X^\top X/n+\lambda I_d)^{-1}$, in the limit $n,d \to \infty$, $d/n\to \infty$. 

As in the isotropic case (Section \ref{sec:mp}), we start by computing a linear pencil for $R$ using \code{auto-fpt}.

\begin{lstlisting}[language=python,numbers=none]
from sympy import Symbol, MatrixSymbol, Identity
import numpy as np
from fpt import calc

# Form the design matrix.
n = Symbol("n", integer=True, positive=True)
d = Symbol("d", integer=True, positive=True)
Z = MatrixSymbol("Z", n, d)
Sigma_sqrt = MatrixSymbol("\Sigma_{sqrt}",d , d)  # Covariance matrix
phi = Symbol(r"\phi", positive=True)

X = Z@Sigma_sqrt  # Design matrix
expr = (X.T@X + Identity(d)).inv()  # Resolvent matrix

# Compute linear pencil using NCAlgebra
Q, (u, v) = compute_minimal_pencil(expr)
\end{lstlisting}
The above code snippet yields:
\begin{align}
Q &= \begin{bmatrix}\mathbb{I} & - Z^\top & 0 & 0\\0 & \mathbb{I} & - Z & 0\\0 & 0 & \mathbb{I} & - \Sigma_{sqrt}\\\Sigma_{sqrt}^\top & 0 & 0 & \mathbb{I}\end{bmatrix},\\
u&=[0, 0, 0, 1],\\
v&=[0, 0, 0, 1].
\end{align}

Similarly to the previous calculation (the isotropic setup), we shall later normalize to take $\lambda$ and the $1/n$ factor into account. We can now compute the fixed-point equations defining the limiting value of $\ntrace R$ like so:

\begin{lstlisting}[language=python,numbers=none]
row_idx, col_idx = np.argmax(u), np.argmax(v)
eqns = calc(Q, random_matrices=["Z"], row_idx=row_idx, col_idx=col_idx,
           normalize="full", subs={d: n * phi})
\end{lstlisting}
This code snippet spits out the following fixed-point equations:
\begin{gather*}
    {G}_{3,3} = \bar{tr}{\left(\lambda \left(\lambda \mathbb{I} + \Sigma_{sqrt}^{2} {G}_{1,1}\right)^{-1} \right)},\\
{G}_{1,1} = - \frac{\lambda}{\phi {G}_{2,0} - \lambda},\\
{G}_{2,0} = \bar{tr}{\left(- \lambda \Sigma_{sqrt}^{2} \left(\lambda \mathbb{I} + \Sigma_{sqrt}^{2} {G}_{1,1}\right)^{-1} \right)}
\end{gather*}
Eliminating $G_{2,0}$ from the above and recalling that $\Sigma_{sqrt}^2=\Sigma$ by construction, we can reorganize these equations as follows:
\begin{align}
    \ntrace R \simeq G_{3,3}/\lambda &= \ntrace (G_{1,1}\Sigma + \lambda I_d)^{-1},\\ \text{ where }
    \lambda/G_{1,1} &= \lambda + \phi \kappa \ntrace\Sigma(\Sigma + (\lambda/G_{1,1}) I_d)^{-1}.
\end{align}
Setting $\kappa:=\lambda/G_{1,1}$, we deduce that $\ntrace R \simeq (\kappa/\lambda)\trace (\Sigma + \kappa I_d)^{-1}$, where $\kappa>0$ uniquely solves the fixed-point equation:
\begin{eqnarray}
    \kappa = \lambda + \phi \kappa \ntrace\Sigma(\Sigma+\kappa I_d)^{-1}.
\end{eqnarray}
This is a well-known result \citep{MP1967} (also see Proposition 1 of \citet{bach2024high} for an alternative derivation).

\section{Subordination with \code{auto-fpt}}
\label{sec:subordination}

\begin{lstlisting}[language=python,numbers=none]
# Input dimension and sample sizes
d = Symbol("d", integer=True, positive=True)
n = Symbol("n", integer=True, positive=True)
lambd = Symbol(r'\lambda', integer=True)
n1 = Symbol("n_1", integer=True, positive=True)  # p_1 * n
n2 = Symbol("n_2", integer=True, positive=True)  # p_2 * n
p1 = Symbol(r"p_1", positive=True)
p2 = Symbol(r"p_2", positive=True)
phi = Symbol(r"\phi", positive=True)

# Design Matrices
Z1 = MatrixSymbol("Z_1", n1, d)
Z2 = MatrixSymbol("Z_2", n2, d)
S1 = MatrixSymbol("S_1", d , d)  # Sqrt of covariance matrix for group 1
S2 = MatrixSymbol("S_2", d , d)  # Sqrt of covariance matrix for group 2
X1 = Z1@S1
X2 = Z2@S2

# Empirical Covariance Matrices
M1 = X1.T@X1
M2 = X2.T@X2
M = M1 + M2

# Compute minimal linear pencil of resolvent R of M using NCAlgebra
R = (M + Identity(d)).inv()
Q, (u, v) = compute_minimal_pencil(R)

# Compute fixed-point equations for expected trace of R
row_idx = np.flatnonzero(u)[0]
col_idx = np.flatnonzero(v)[0]

variances = {Z1: 1 / (n * lambd), Z2: 1 / (n * lambd)}
eqns = calc(Q, random_matrices=["Z_1", "Z_2"], row_idx=row_idx,
           col_idx=col_idx, variances=variances, subs={d: n * phi})
\end{lstlisting}

\section{Random Features Model with \code{auto-fpt}}
\label{sec:rf-model}

Equations \ref{eq:y-pred-ntk} to \ref{eq:tau1-tau2-rat} are largely from \citet{adlam2020neural}, which we include here for completeness.
The predictions $\hat{y}(X)$ of the network on the training set (when trained using gradient descent and the width of the network becomes large) converges to:
\begin{align}
    \hat{y} (x) &= N_0 (X) + (Y - N_0 (X)) K^{-1} K(X, X) \\
    &= Y - \gamma (Y - N_0 (X)) K^{-1}.
    \label{eq:y-pred-ntk}
\end{align}
We make use of the following definitions:
\begin{gather*}
    K := K(X, X) + \lambda I_n, K(x_1, x_2) = J(x_1) J(x_2)^\top,
\end{gather*}
where $\lambda$ is the regularization penalty and $J$ is the Jacobian of the network's output with respect to its parameters. In the setting $\sigma^2_{W_2} \to 0$, because $K_1 (X, X) = (\partial N_0 (X) / \partial W_1) (\partial N_0 (X) / \partial W_1)^\top = 0$, we have that:
\begin{align}
K(X, X) &= K_2 (X, X) = (\partial N_0 (X) / \partial W_2) (\partial N_0 (X) / \partial W_2)^\top \\
&= F^\top F / m,
\end{align}
where the feature matrix $F:=\sigma(W_1X/\sqrt d)$. Then, the training risk can be decomposed as follows:
\begin{align}
    E_{train} &= \mathbb E \ntrace ((Y - \hat{y} (X))^\top (Y - \hat{y} (X))) \\
    &= \gamma^2 \mathbb E \ntrace ((Y - N_0 (X))^\top (Y - N_0 (X)) K^{-2}) \\
    &= T_1 + T_2, \\
    \text{ with } T_1 &= \gamma^2 \mathbb E \ntrace Y^\top Y K^{-2}, \\
    T_2 &= \gamma^2 \mathbb E \ntrace N_0 (X)^\top N_0(X) K^{-2},
\end{align}
where we have used that the expectation of terms linear in $N_0(X)$ is 0. We observe that because $\sigma_{W_2}^2 \to 0$:
\begin{align}
    N_0 (X) &= W_2 F / \sqrt{m}, \\
    T_2 &= \mathbb E \ntrace W_2^\top W_2 (F / \sqrt{m}) (F^\top F / m)^{-2} (F / \sqrt{m})^\top \\
    &= \sigma_{W_2}^2 \mathbb E \ntrace (F / \sqrt{m}) (F^\top F / m)^{-2} (F / \sqrt{m})^\top = 0.
\end{align}

Now, the \emph{Gaussian Equivalence Theorem} \citep{goldt2022gaussian} tells us that we can replace $F$ with a linearized version $F_{lin}$, and $Y$ similarly with $Y_{lin}$:
\begin{align}
F_{lin}&=\sqrt{\frac{\zeta}{d}}W_1 X+\sqrt{\beta}\Theta_F,\quad Y_{lin}=\sqrt{\frac{1}{d_T d}} \omega \Omega X+ \epsilon,
\end{align}
without changing their limiting spectral distribution.
Here, $\Theta_F$ is a random $m \times n$ matrix independent of $X$ and $W_1$, with IID entries from $\mathcal N(0,1)$. Hence, we deduce that:
\begin{align}
    T_1 &= \gamma^2 \mathbb E \ntrace Y_{lin}^\top Y_{lin} K^{-2} \\
    &= \gamma^2 \mathbb E \ntrace (X^\top X / d + \sigma_\epsilon^2 I_n) K^{-2} \\
    &= \gamma^2 (\mathbb E \ntrace (X^\top X K^{-2}) / d + \sigma_\epsilon^2 \mathbb E \ntrace K^{-2}) \\
    &= -\gamma^2 (\tau_2' + \sigma_\epsilon^2 \tau_1'), \\
\end{align}
where $\tau_1(\lambda) := \mathbb E \ntrace K^{-1}$ and $\tau_2(\lambda) := \frac{1}{d} \mathbb E \ntrace X^\top X K^{-1}$. It now remains to derive fixed-point equations relating $\tau_1, \tau_2$. We define the following random matrices:
\begin{gather*}
    W_0 = \sqrt{\frac{\zeta}{\lambda m}} W_1,\quad X_0 = \sqrt{\frac{1}{d}} X, \quad \Theta_0 = \sqrt{\frac{\beta}{\lambda m}} \Theta_F, \\
    F_0 = W_0 X + \Theta_0.
\end{gather*}

Using the above definitions, we can rewrite $\tau_1, \tau_2$ as:
\begin{align}
   \tau_1 &= \frac{1}{\lambda} \mathbb E \ntrace (F_0^\top F_0 + I_n)^{-1}, \\
   \tau_2 &= \frac{1}{\lambda} \mathbb E \ntrace X_0^\top X_0 (F_0^\top F_0 + I_n)^{-1}.
   \label{eq:tau1-tau2-rat}
\end{align}

NCAlgebra can be used to compute a single minimal pencil for both $\tau_1$ and $\tau_2$, and we apply \code{auto-fpt} to derive a reduced system of fixed-point equations relating the quantities.

\begin{lstlisting}
from sympy import Symbol, MatrixSymbol, Identity, pprint, latex, \
    Eq, solve, symbols, fraction, simplify, expand, factor, together
import numpy as np
from fpt import calc

# Define the relevant random and deterministic matrices and scalars.
n = Symbol("n", integer=True, positive=True)
d = Symbol("d", integer=True, positive=True)
m = Symbol("m", integer=True, positive=True)

phi = Symbol(r"\phi", positive=True)
psi = Symbol(r"\psi", positive=True)
lambd = Symbol(r"\lambda", positive=True)
d = phi * n
m = d / psi

zeta = Symbol(r"\zeta", positive=True)
eta = Symbol(r"\eta", positive=True)
beta = Symbol(r"\beta", positive=True)

W0 = MatrixSymbol("W_0", m, d)
X = MatrixSymbol("X", d, n)
T0 = MatrixSymbol(r"\Theta_0", m, n)
variances = {X: 1 / d, W0: zeta / (m * lambd), T0: beta / (m * lambd)}

# Form the resolvent for Kinv.
F0 = W0 @ X + T0
Kinv = (F0.T @ F0 + Identity(n)).inv()

# Compute a minimal linear pencil for X^T * X * Kinv + Kinv using NCAlgebra.
Q, (u, v) = compute_minimal_pencil(X.T @ X @ Kinv + Kinv)

# Get the index of the entry in u, v corresponding to X^T * X * Kinv.
XTXKinv_i = np.flatnonzero(u)[0]
XTXKinv_j = np.flatnonzero(v)[0]

# Compute the fixed-point equations for Kinv and X^T * X * Kinv 
# using free probability theory.
eqns = calc(Q, random_matrices=[X, W0, T0], row_idx=XTXKinv_i, \
            col_idx=XTXKinv_j, subs={}, variances=variances)

# Get the index of the entry in u, v corresponding to Kinv.
Kinv_i = np.flatnonzero(u)[1]
Kinv_j = np.flatnonzero(v)[0]
\end{lstlisting}

The above code snippet yields the following fixed-point equations for $\tau_1, \tau_2$:
\begin{gather*}
    \lambda \tau_2 = \frac{\lambda {G}_{0,0}}{\beta {G}_{4,4} + \lambda {G}_{3,0} + \lambda},\\
{G}_{3,0} = \frac{\phi \zeta {G}_{4,4}}{\lambda \phi + \zeta {G}_{2,2} {G}_{4,4}},\\
\lambda \tau_1 = \frac{\lambda}{\beta {G}_{4,4} + \lambda {G}_{3,0} + \lambda},\\
{G}_{4,4} = - \frac{\lambda \phi}{- \beta \psi {G}_{2,2} - \lambda \phi + \phi \psi \zeta {G}_{0,3}},\\
{G}_{0,3} = - \frac{\lambda {G}_{2,2}}{\lambda \phi + \zeta {G}_{2,2} {G}_{4,4}},\\
{G}_{0,0} = \frac{\lambda \phi}{\lambda \phi + \zeta {G}_{2,2} {G}_{4,4}}.
\end{gather*}

Via variable elimination (e.g., through programmatic iterative substitution), we arrive at the following reduced system relating $\tau_1, \tau_2$:
\begin{align}
    0 &= - \phi \left(\beta \phi \tau_{1}^{2} - \beta \phi \tau_{1} \tau_{2} + \lambda \tau_{1}^{2} \tau_{2} \zeta + \phi \tau_{1} \tau_{2} \zeta - \phi \tau_{2}^{2} \zeta - \tau_{1} \tau_{2} \zeta\right) := p_1, \\
    0 &= - \lambda \phi \left(\beta \psi \tau_{1}^{2} - \beta \psi \tau_{1} \tau_{2} + \phi \tau_{1} - \phi \tau_{2} + \psi \tau_{1} \tau_{2} \zeta - \psi \tau_{2}^{2} \zeta - \tau_{1} \tau_{2} \zeta\right) := p_2.
\end{align}

By substituting $\beta = \eta - \zeta$ and dividing out constant factors, we observe that:
\begin{align}
    0 &= p_1 / \phi = - \eta \phi \tau_{1}^{2} + \eta \phi \tau_{1} \tau_{2} - \lambda \tau_{1}^{2} \tau_{2} \zeta + \phi \tau_{1}^{2} \zeta - 2 \phi \tau_{1} \tau_{2} \zeta + \phi \tau_{2}^{2} \zeta + \tau_{1} \tau_{2} \zeta \text{ (S68 \citep{adlam2020neural})}, \\
    0 &= -\psi p_1 / \phi + p_2 / \lambda = \lambda \psi \tau_{1}^{2} \tau_{2} \zeta - \phi^{2} \tau_{1} + \phi^{2} \tau_{2} + \phi \tau_{1} \tau_{2} \zeta - \psi \tau_{1} \tau_{2} \zeta \text{ (S67 \citep{adlam2020neural})}.
\end{align}

This exactly recovers the coupled polynomial equations for $\tau_1, \tau_2$ in \citet{adlam2020neural}, namely Proposition 1 thereof with $\sigma^2_{W_2} = 0$. We provide code for programmatic empirical and analytical validation of the equations for $\tau_1, \tau_2$ under the \code{examples} folder in our repository.

\end{document}